\documentclass{article}

\usepackage{arxiv}

\usepackage[utf8]{inputenc} 
\usepackage[T1]{fontenc}    
\usepackage{hyperref}       
\usepackage{url}            
\usepackage{booktabs}       
\usepackage{amsfonts}       
\usepackage{nicefrac}       
\usepackage{microtype}      
\usepackage{lipsum}
\usepackage{bbold}

\usepackage{mathrsfs}
\usepackage{amsmath}
\usepackage{amssymb}
\usepackage{graphicx}

\title{
Robust computation with rhythmic spike patterns
}

\author{{\large \bf E. Paxon Frady (epaxon@berkeley.edu)} 
\AND {\large \bf Friedrich T. Sommer (fsommer@berkeley.edu)}\\
Redwood Center, UC Berkeley, \\ 
Berkeley, CA 94720 USA
}

\begin{document}
\maketitle

\begin{abstract}
Information coding by precise timing of spikes can be faster and more energy-efficient than traditional rate coding. 
However, spike-timing codes are often brittle, which has limited their use in theoretical neuroscience and computing applications. Here, we propose a novel type of attractor neural network in complex state space, and show how it can be leveraged to construct spiking neural networks with robust computational properties through a phase-to-timing mapping.
Building on Hebbian neural associative memories, like Hopfield networks, we first propose threshold phasor associative memory (TPAM) networks. Complex phasor patterns whose components can assume continuous-valued phase angles and binary magnitudes can be stored and retrieved as stable fixed points in the network dynamics. 
TPAM achieves high memory capacity when storing sparse phasor patterns, and we derive the energy function that governs its fixed point attractor dynamics.  
Second, through simulation experiments we show how the complex algebraic computations in TPAM can be approximated by a biologically plausible network of integrate-and-fire neurons with synaptic delays and recurrently connected inhibitory interneurons. 
The fixed points of TPAM in the complex domain are commensurate with stable periodic states of precisely timed spiking activity that are robust to perturbation.
The link established between rhythmic firing patterns and complex attractor dynamics has implications for the interpretation of spike patterns seen in neuroscience, and can serve as a framework for computation in emerging neuromorphic devices.
\end{abstract}

\section*{Introduction}

The predominant view held in neuroscience today is that the activity of neurons in the brain and the function of neural circuits can be understood in terms of the computations that underlie perception, motion control, decision making, and cognitive reasoning. 
However, recent developments in experimental neuroscience have exposed critical holes in our theoretical understanding of how neural dynamics relates to computation. 
For example, in the prominent existing theories for neural computation, such as energy-based attractor networks \citep{Hopfield1984, amit1992book}, population coding \citep{Pouget2003} and the neural engineering framework \citep{Eliasmith2012}, information is represented in firing rates of neurons, as opposed to precise timing patterns of action-potentials. The rate-coding assumption \citep{Bialek1991, Cessac2010} is hard to reconcile with dynamic phenomena ubiquitously observed in neural recordings, such as precise sequences of action-potentials \citep{Abeles1982,Reinagel2000, Panzeri2010, Tiesinga2008}, high speed of neural computation during perception \citep{Thorpe1996}, as well as rhythmic activity of neural populations on various temporal and spatial scales \citep{OKeefe1993,Buzsaki2004}.   

Here, we develop a theory that can elucidate the role of precise spike timing in neural circuits. The theory is based on complex-valued neural networks and a \emph{phase-to-timing} mapping that translates a complex neural state to the
timing of a spike. We first introduce a novel model of fixed point attractor networks in complex state space, called \emph{threshold phasor associative memory} (TPAM) networks. The model dynamics is governed by a Lyapunov function, akin to Hopfield networks \citep{Hopfield1982}, it has high memory capacity and, unlike Hopfield networks, can store continuous-valued data. Second, we show that any TPAM network possesses a corresponding equivalent network with spiking neurons and time-delayed synapses. Our framework can be used to design circuits of spiking neurons to compute robustly with spike times, potentially trailblazing a path towards fully leveraging recent high-performance neuromorphic computing hardware \citep{Davies2018}. Concurrently, the framework can help connect computations with experimental observations in neuroscience, such as
sequential and rhythmic firing dynamics, balance between excitation and inhibition, and synaptic delays.

\section*{Background}
\label{sec:background}

Fixed point attractor models, and more generally, energy-based models \citep{ackley1985learning}, have played an extremely important role in neural networks and theoretical neuroscience. The appeal of these models comes from the fact that their dynamics can be described by the descent of an Energy or Lyapunov function, often conceptualized as an ``energy landscape''.  
While energy descent does not describe all of neural computation, it is the basis of neural network algorithms for many important problems, such as denoising/error correction to make computations robust with respect to perturbations, or constrained optimization \citep{Hopfield1985}.

\subsection*{The landscape of fixed point attractor neural networks}
\label{sec:attractor}
\begin{figure}[ht]
\centering
\includegraphics[width=0.3\textwidth]{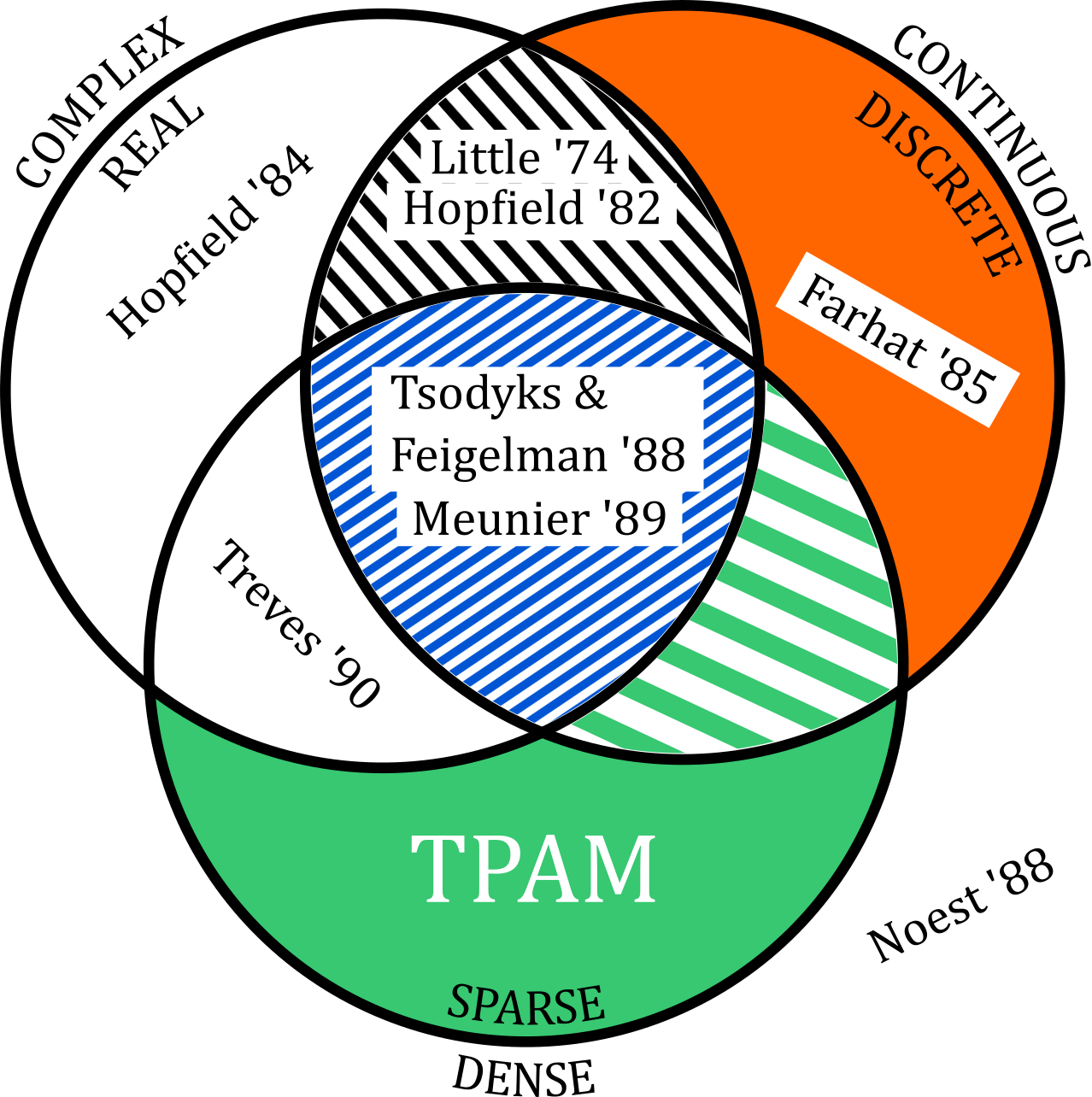}
\caption{Venn diagram delineating different types of fixed point attractor networks. The boundaries represent the following distinctions: 1) complex-valued versus real-valued neural states, 2) continuous-valued versus discrete-valued neural states, and 3) dense versus sparse neural activity patterns. Regions are labeled by key publications describing these models: 
Little '74 = \citep{little1974existence}, Hopfield '82 = \citep{Hopfield1982}, Hopfield '84 = \citep{Hopfield1984}, Farhat '85 = \citep{farhat1985optical}, Tsodyks \& Feigelman '88 = \citep{tsodyksfeigelman1988enhanced}, Meunier '89 = \citep{meunieretal1989threestate}, Treves '90 = \citep{treves1990graded}, Noest '88 = \citep{noest1988phasor}.
Our research focuses on the uncharted areas, sparse complex-valued models (TPAM). 
}
\label{fig:attractor_landscape}
\end{figure}

Here, we focus on a prominent class of fixed point attractor networks with Hebbian one-shot learning to store a set of neural activity patterns. To retrieve a pattern, the network is first initialized with a cue -- typically a noisy or partial version of a stored pattern. For retrieval, an iterative dynamics successively reduces the initial noise and converges to a clean version of the stored memory. The models can be distinguished along the following dimensions: real-valued versus a complex-valued neural state space, neurons with discretizing versus continuous transfer functions, and the neural firing patterns in the network being sparse versus dense (Fig. \ref{fig:attractor_landscape}).

The traditional models of attractor neural networks, which have a symmetric synaptic matrix that guarantees Lyapunov dynamics and fixed points \citep{cohengrossberg1983absolute}, were inspired by the Ising model of ferromagnetism \citep{ising1925beitrag}. The neural transfer function is step-shaped (describing the spiking versus silent state of a neuron), and the models can store dense activity patterns with even ratio between active and silent neurons \citep{little1974existence, Hopfield1982}. The inclusion of ``thermal'' noise yielded networks with a sigmoid-shaped neural transfer function \citep{Hopfield1984,treves1990graded,kuhnetal1991statistical}, which produces continuous real-valued state vectors that could be conceptually related to firing rates of spiking neurons in the brain. Further, following the insight of Willshaw et al. \citep{willshaw1969non}, attractor networks for storing sparse activity patterns were proposed, which better matched biology and exhibited greatly enhanced memory capacity \citep{tsodyksfeigelman1988enhanced,buhmannetal1989associative,palmsommer1992information,schwenker1996iterative}.

One drawback of these traditional memory models is that all attractor states are (close to) binary patterns, where each neuron is either almost silent or fully active near saturation.
The restriction to binary memories can be overcome by introducing model neurons that can saturate at multiple (more than two) activation levels \citep{gross1985mean,meunieretal1989threestate,yedidia1989neural,bouten1993basin}. This class of models was inspired by the Potts glass model in solid state physics. Another model with multi-level neurons is the so-called {\it complex Hopfield network} \citep{farhat1985optical,cook1989mean,gerl1992learning,Hirose1992,Jankowski1996,Donq-Liang1998,Aoki2000,Aizenberg2011we,Kobayashi2017}. Here, the model neurons are discretized phasors, and as a result, the states of the model are complex vectors whose components have unity norm and phase angles chosen from a finite, equidistant set. For a discretization number of two, complex Hopfield networks degenerate to the real-valued bipolar Hopfield network \citep{Hopfield1982}. 

A unique strength of a complex state space was highlighted by {\it phasor networks}  \citep{noest1988phasor,sirat1989grey, Hoppensteadt1996}. In phasor networks, the neural transfer function is a phasor projection, i. e. each neural state carries the continuous phase value of the postsynaptic sum, but has normalized magnitude. Interestingly, even without phase discretization, phasor networks can store arbitrary continuous-valued phasor patterns. Patterns with arbitrary relative phase angles can be stable because of the ring topology of normalized complex numbers. 

In existing phasor networks and complex Hopfield networks, all neurons represent phases at every time step. To provide models of greater computational versatility, here we explore relatively uncharted territory in Figure~\ref{fig:attractor_landscape}: Attractor networks with complex-valued sparse activity states. These models can also store phasor patterns, in which a fraction of components have zero amplitude that correspond to silent or inactive neurons. Specifically, we introduce and investigate a novel attractor network model called \emph{threshold phasor associative memory} (TPAM) network. As will be shown, pattern sparsity in TPAM enables high memory capacity -- as in real-valued models \citep{tsodyksfeigelman1988enhanced}, and also corresponds to spike timing patterns that are neurobiologically plausible.

\subsection*{Hebbian sequence associative memories}
\label{sec:sequence_models}
A first foray into temporal neural coding was the development of networks of threshold neurons with Hebbian-type hetero-associative learning that synaptically stores the first-order Markovian transitions of sequential neural activity patterns \citep{Amari1972, Willwacher1976, Willwacher1982,Kleinfeld1986,Sompolinsky1986,Amit1988,riedeletal1988temporal}. When initialized at or near the first pattern of a stored sequence, the parallel-update and discrete time dynamics of the network will produce the entire sequence, with a pattern transition occurring each time step. These networks have non-symmetric synaptic matrices and therefore the dynamics are not governed by a Lyapunov function  \citep{cohengrossberg1983absolute}. 
However, for networks that store cyclic pattern sequences of equal lengths, Herz et al. \citep{herz1991global, herzetal1991statistical} have shown that the network dynamics are governed by an energy function, defined in an extended state space in which each entire sequence is represented by a point. 

Our approach, to describe temporal structure in the complex domain, is related to the extended state space approach. Specifically, we  will show that sequence associative networks correspond to attractor networks whose fixed points are phase patterns with equidistantly binned phase values, each phase bin representing a position in the sequence.  

\subsection*{Theories of spiking neural networks}
The firing rates in a spiking neural network can be described by attractor networks with sigmoid neural transfer function \citep{Hopfield1984,treves1990graded,kuhnetal1991statistical}. This modeling approach rests on the paradigm of rate coding \citep{Bialek1991, Cessac2010}, which considers spiking as an inhomogenious Poisson random process, with synaptic inputs controlling only the instantaneous event rate.  

Around 1990, the development of theory to describe networks of neurons with explicit spiking mechanisms began.
Using phase-return maps, Mirollo and Strogatz \citep{mirollo1990synchronization} found that integrate-and-fire neurons coupled without synaptic delays synchronize for almost all initial conditions. 
Such networks with delay-less synaptic couplings formed by outer-product learning rules have been demonstrated to serve as autoassociative memories. Depending on the operation point, retrieval states can either be represented by synchronious spike patterns \citep{buhmann1986associative,muellerherz1999content,sommer2001associative}, or by asynchroneous rate patterns \citep{gerstnervanhemmen1992associative}. 
Further it was shown that for certain types of delay-less synaptic matrices, there are stable limit cycle attractors in which every neuron fires once per cycle \citep{hopfieldherz1995rapid}. The convergence to these dynamic attractor states is rapid and can be described by the descent in a Lyapunov function.  
   
Here, we are particularly interested in networks in which spiking neurons are coupled by synapses, each with individual delay time, whose properties have been studied in small networks \citep{ernstetal1995synchronization,nischwitz1995local}. 
Herz \citep{herzetal1995preprint} derived a Lyapunov function for networks of non-leaky integrate-and-fire neurons with synaptic delays. As in the delay-less case \citep{hopfieldherz1995rapid}, the descent in the Lyapunov function corresponds to the rapid (although not very robust) convergence to particular temporal coding patterns. Somewhat inspired by \citep{herzetal1995preprint}, our novel approach is to map complex fixed point attractor networks to networks of leaky integrate-and-fire neurons. 
As we will show, the resulting spiking neural networks employ spike timing codes that are robust with regard to perturbations, exhibit high memory capacity, and share many properties with neuronal circuitry of real brains.

\section*{Results}

\subsection*{Threshold phasor associative memories}
\label{section:aaTPAM}

We propose a novel memory model, the threshold phasor associative memory (TPAM), which can store sparse patterns of complex phasors as fixed point attractors. The network dynamics is governed by an energy function, which we derive. Further, simulation experiments show that TPAM has high memory capacity and provides an efficient error correction stage in a neural network for storing images.

\subsubsection*{Learning and neural dynamics}

Phasor neural networks \citep{noest1988phasor,sirat1989grey} were designed to store dense phase angle patterns in which every component represents a phase angle. Similar to auto-associative outer product Hebbian learning \citep{kohonen1972correlation}, like in the standard Hopfield network \citep{Hopfield1982}, phasor networks employ a complex-conjugate outer-product learning rule:
\begin{equation}
\mathbf{W} = \mathbf{S} \mathbf{S}^{*\top} \;\; , \; W_{ij} = \sum_{m=1}^M r_{im} r_{jm} e ^ {\textrm{i} (\phi_{im} - \phi_{jm})} 
\label{phasor_learning}
\end{equation}
where $\mathbf{S} \in \mathbb{C}^{N \times M}$ is a matrix of $M$ phasor patterns of dimension $N$. A component of one of the stored patterns is given by $S_{im} = r_{im} e ^ {\textrm{i} \phi_{im}}$. The entries along the diagonal of $\mathbf{W}$ are set to $0$. 

During the discrete-time neural update in a phasor network, the output of neuron $j$ is a unit magnitude complex number $z_j$. The complex neural states are multiplied by the complex weight matrix to give the postsynaptic dendritic sum: 
\begin{equation}
u_i(t) = \sum_j W_{ij} z_j(t)
\label{dend_sum}
\end{equation}

In contrast to the described phasor memory network, the TPAM network is designed for storing patterns in which only a sparse fraction of components $p_{hot} = K/N$ have unit magnitude and the rest have zero amplitude, i.e., are inactive. TPAM uses the same learning rule [\ref{phasor_learning}] and postsynaptic summation [\ref{dend_sum}] as the original phasor network, but differs in the neural transfer function. The neural transfer function includes a threshold operation on the amplitude of the synaptic sum [\ref{dend_sum}]: 
\begin{equation}
 z_i(t+1) = g(u_i(t), \Theta(t)) := \frac{u_i(t)}{|u_i(t)|} H(|u_i(t)|-\Theta(t))
\label{neural_transfer}
\end{equation}
with $H(x)$ the Heaviside function.
If the threshold $\Theta(t)$ is met, the output preserves the phase of the sum vector and normalizes the amplitude. Otherwise, the output is zero.

To maintain a given level of network activation, the threshold setting needs to be controlled as a function of the global network activity \citep{wennekersetal1995threshold}.
Here, we set the threshold proportional to the overall activity:
\begin{equation}
    \Theta(t) = \theta \sum_i |z_i(t)| = \theta |\mathbf{z}(t)|
    \label{eqn:threshold_strat}
\end{equation}
with $\theta$ a scalar between 0 and 1, typically slightly less than 1.

\begin{figure}[ht]
    \centering
    \includegraphics[width=0.65\textwidth]{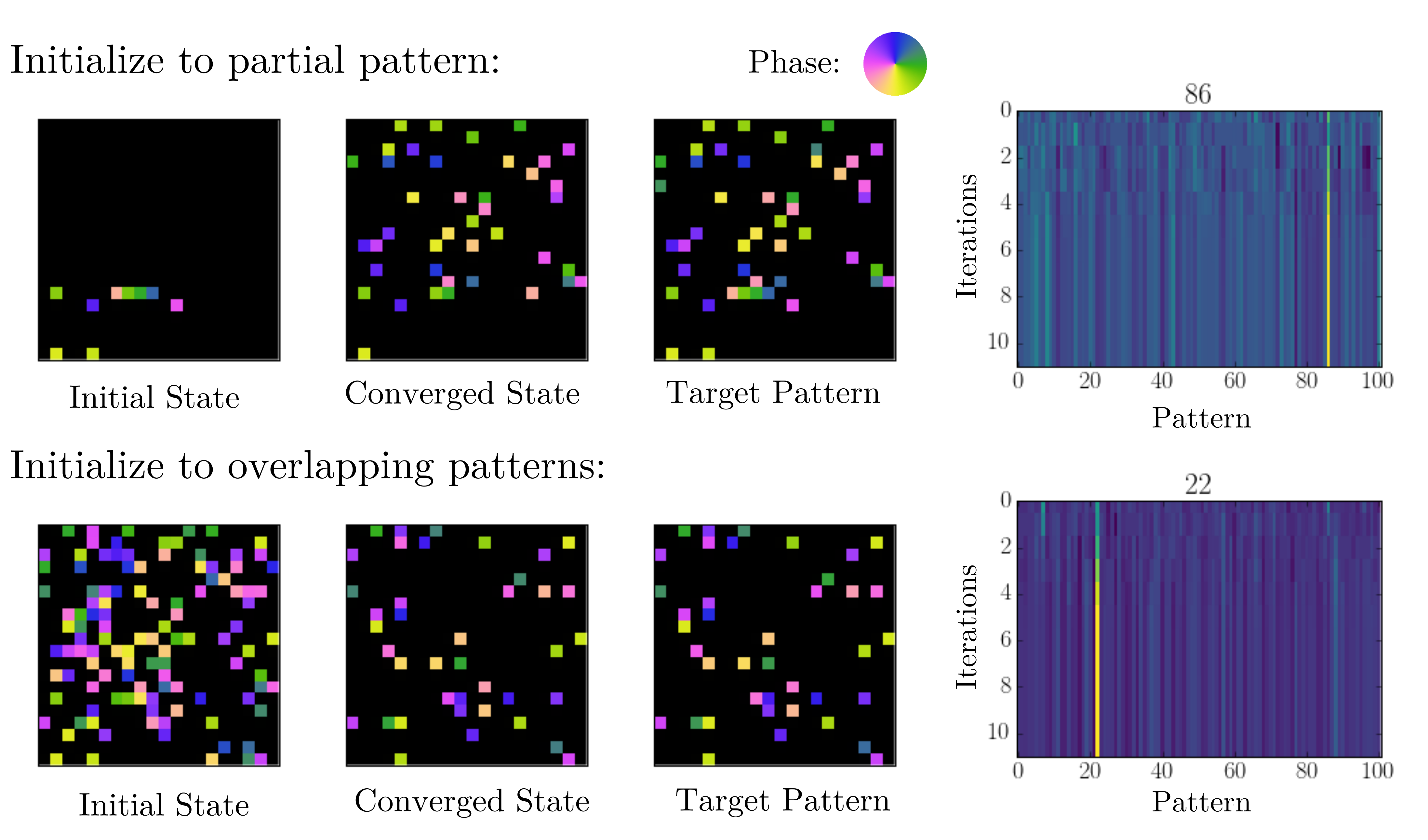}
    \caption{\textbf{Memory recall in a TPAM network.} Results of two retrieval experiments (parameters see text), one initialized by a partial memory pattern (top row), one by a superposition of three memory patterns (bottom row). Both recalls were successful, as indicated by the similarity between ``converged'' and ``target'' patterns (phase values color coded; black corresponds to zero amplitude). Panels on the right show that it takes only a few iteration steps until only the overlap with the retrieved memory is high (yellow).}
    \label{fig:sparse_chop}
\end{figure}

The memory recall in TPAM with parallel update with $400$ neurons is demonstrated in Fig. \ref{fig:sparse_chop}.
The network has stored $100$ sparse random phasor patterns with $p_{hot}=10\%$ and phase values drawn independently from a uniform distribution.
The iterative recall is initialized by a partial memory pattern -- with some nonzero components set to zero (top panels), and with a superposition of several stored patterns (bottom panels). In both cases, the network dynamics relaxes to one of the stored memories (approximately).

\subsubsection*{Energy function of TPAM networks}

For traditional phasor memory networks (without threshold), Noest \citep{noest1988phasor} showed that the corresponding Lyapunov function is
\begin{equation}
E(\mathbf{z}) = - \frac{1}{2} \sum_{ij} W_{ij} z_i z_j^*
\label{phasor_lyapunov}
\end{equation}
Note, that because [\ref{phasor_learning}] results in a Hermitian matrix $\mathbf{W}$, [\ref{phasor_lyapunov}] is a real-valued function. Further note, that the dynamics in phasor networks is a generalization of phase-coupled systems well studied in physics, such as the Kuramoto model \citep{kuramoto1975self}, and the XY model \citep{kosterlitz1973ordering}, and for describing coherent activity in neural networks \citep{schuster1990model,sompolinsky1990global,niebur1991oscillator}. Those models are governed by a Lyapunov function of the form [\ref{phasor_lyapunov}], but in which $\mathbf{W}$ is real-valued and symmetric \citep{vanhemmenwreszinski1993lyapunov}.

To see how the inclusion of the threshold operation in the TPAM update [\ref{neural_transfer}] changes the Lyapunov function, we follow the treatment in \citep{Hopfield1984} by extending [\ref{phasor_lyapunov}] to describe the dynamics of phasor networks with arbitrary invertible transfer function $f(z)$: 
\begin{equation}
E(\mathbf{z}) = - \frac{1}{2}\sum_{ij} W_{ij} z_i z_j^* + \sum_i \int_0^{|z_i|} f^{-1}(v) dv
\label{lyapunov}
\end{equation}

The neural transfer function of TPAM, $g(z;\Theta)$ in [\ref{neural_transfer}], is not invertible. But it can be approximated by a smooth, invertible function by replacing the Heaviside function in [\ref{neural_transfer}] with an invertible function $f(z)$, for example, the logistic function.  
In the limit of making the approximation tight, i.e., $f(z)\approx g(z;\Theta)$, the corresponding update is given by [\ref{neural_transfer}]. For a constant global threshold $\Theta = \Theta(t)$ the Lyapunov function [\ref{lyapunov}) of TPAM is:
\begin{equation}
E(\mathbf{z}) = - \frac{1}{2}\sum_{ij} W_{ij} z_i z_j^* + \Theta|\mathbf{z}| +b(\mathbf{z}) 
\label{lyapunov_tpan}
\end{equation}
with a potential barrier function $b(\mathbf{z})=\infty \; (1- \prod_i H(1-|z_i|))$. According to equation [\ref{lyapunov_tpan}], a positive constant global threshold [\ref{neural_transfer}] has the effect of adding a $L_1$ constraint term, which encourages a lower activity in the network.

For the dynamic threshold control [\ref{eqn:threshold_strat}], the Lyapunov function for TPAM becomes:
\begin{equation}
E(\mathbf{z}) = \sum_{ij} \left( - \frac{1}{2} W_{ij} + \theta \mathbb{I} \right) z_i z_j^* +b(\mathbf{z}) 
\label{lyapunov_tpan_dyn}
\end{equation}
According to equation [\ref{lyapunov_tpan_dyn}], a positve coefficient $\theta$ in the dynamic threshold control,  [\ref{neural_transfer}] and [\ref{eqn:threshold_strat}], adds a repulsive self-interaction between active phasors, thereby reducing the activity in the network.

The derived Lyapunov functions help to clarify the difference between constant and linear threshold control. Consider the case of low memory load. With constant threshold, not only are the individual stored patterns stable fixed points, but also their superpositions will be stable. In contrast, dynamic threshold control introduces competition between active stored memory patterns. The coefficient $\theta$ can be tuned so that only individual patterns are stable (as done here). When lowered, superpositions of two (or more) patterns can become stable, but competition still only allows a limited number of active superpositions. This may be useful behavior for applications outside the scope of this paper.

\subsubsection*{Information capacity of TPAM networks}

To understand the function of TPAM, the impact of its different features on memory performance are studied through simulation experiments. After storing $M$ random patterns, we initialize the network to one of the stored patterns with a small amount of noise. The network runs until convergence or for a maximum of 500 iterations. To assess the quality of memory recall we then compare the network state with the error-less stored pattern. 

\begin{figure}[ht]
    \centering
    \includegraphics[width=0.5\textwidth]{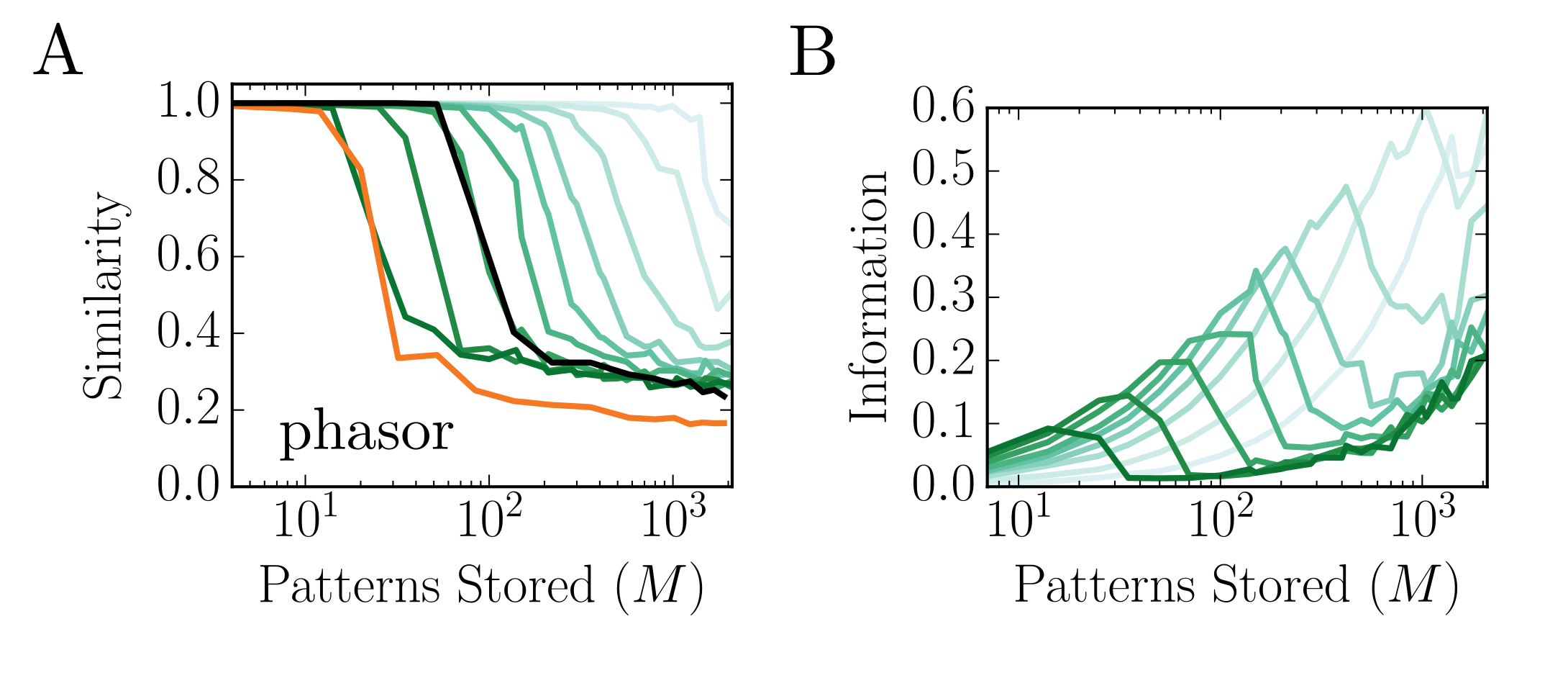}
    \caption{\textbf{Capacity of TPAM networks.}
    A. Recall performance of TPAM as a function of stored patterns (green; lighter green indicates higher sparsity), in comparision with traditional associative memory models: bipolar Hopfield networks (black) \citep{Hopfield1982}, and continuous phasor networks (orange) \citep{noest1988phasor}.  Similarity is the average cosine of the angle between components with nonzero magnitudes in 
    converged network state and target state. 
    B. The memory capacity of the TPAM network in bits per synapse (sparsity encoded by shading, as in panel A).
}
    \label{fig:tpam_capacity}
\end{figure}

Figure \ref{fig:tpam_capacity}A displays on the $y$-axes \emph{cosine similarity} (i.e. correlation) between the output of the memory and the desired target pattern. This normalized metric accounts for both disparity in the phase offset and mismatch in the supports, but does not directly reflect the mutual information between patterns, which also depends on the sparsity level.
Fig. \ref{fig:tpam_capacity}A compares TPAM with different levels of sparsity (green) to the traditional binary Hopfield network \citep{Hopfield1982} (black line), and to the continuous phasor network \citep{noest1988phasor} (orange line).
As in the case of the ternary models \citep{meunieretal1989threestate} (see Supplement), the number of patterns that can be recalled with high precision increases significantly with sparsity.  

To assess how the memory efficiency of TPAM depends on pattern sparsity, we empirically measure the information in the random phase patterns that can be recalled from the memory (see Supplement). 
Dividing the recalled information by the number of synapses yields the {\it memory capacity} of a network in bits per synapse \citep{palmsommer1992information, schwenker1996iterative}. Measuring the memory capacity in bits, rather than by the number of stored patterns \citep{Hopfield1982}, has the advantage that performances can be compared for memory patterns with different sparsity levels.

The measurements of memory capacity in bits/synapse shows that sparsity greatly increases memory capacity (Fig. \ref{fig:tpam_capacity}B) over dense associative memory networks. 
Interestingly, this result parallels the increase of memory capacity in binary Hopfield networks with pattern sparsity \citep{tsodyksfeigelman1988enhanced, buhmannetal1989associative, palmsommer1992information}. 
This holds up to a limit, however, as the networks with the highest sparsity levels had a slightly decreased maximum memory capacity in the simulation experiments. 

\subsubsection*{Indexing and retrieving data with TPAM networks}
One option to store real-world data in TPAM networks is to encode the data in phasor patterns that can be stored in a recurrent TPAM network as described in the last section. A problem with this approach is that data correlations cause interference in the stored patterns, which is a known issue in traditional associative memories with outer product learning rule that reduces the information capacity quite drastically \citep{amit1992book}.  

\begin{figure}[ht]
    \centering
    \includegraphics[width=0.65\textwidth]{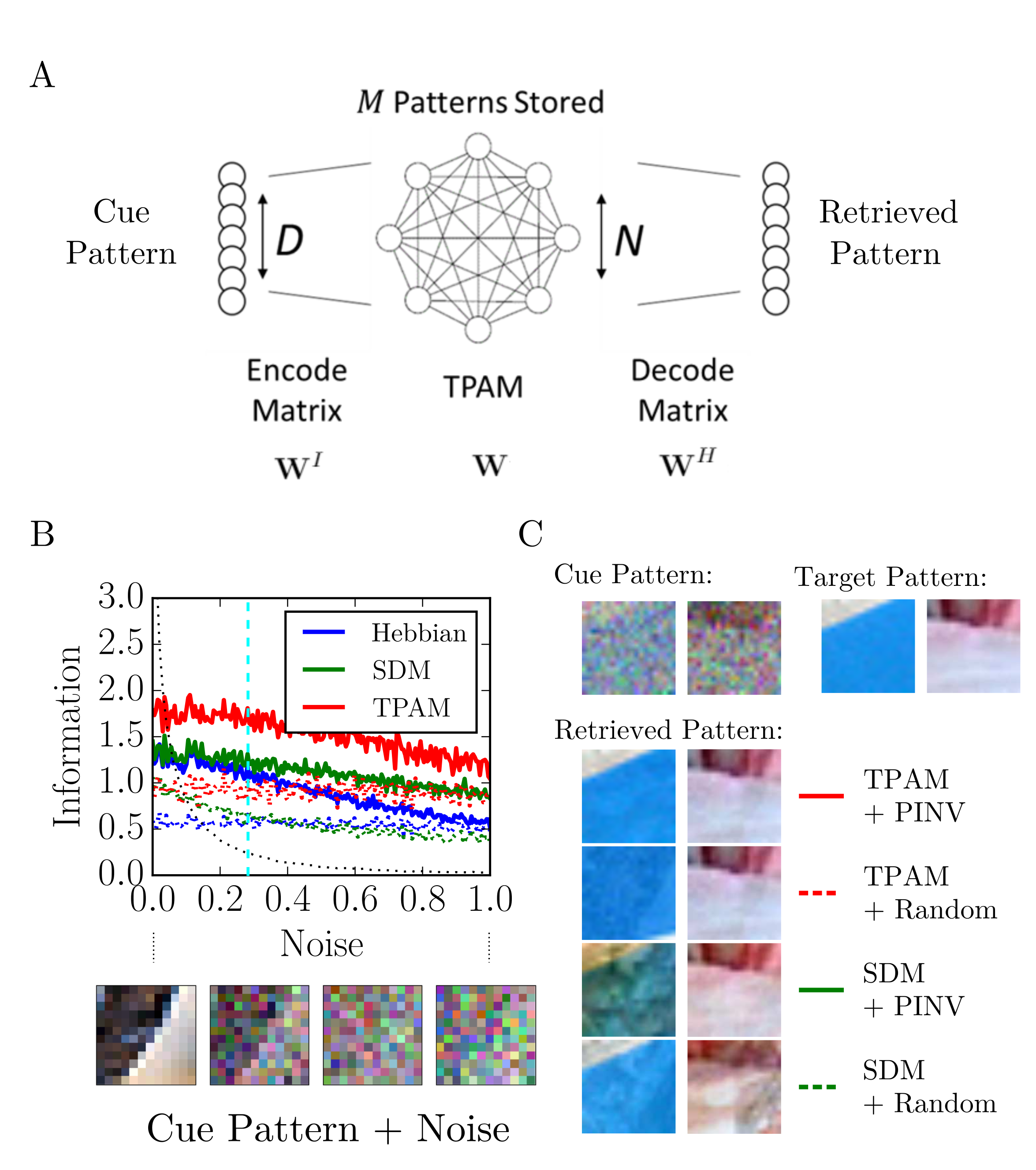}
    \caption{\textbf{Storage of natural images in a three layer architecture with complex neurons}. A. Layered network architecture, consisting of three stages, indexing stage with pattern separation, TPAM for error correction, and heteroassociative memory. B. Comparision of the performances of the full architecture versus other heteroassociative memory models. The pattern separation stage for indexing (solid lines) and the TPAM network for error correction significantly increases the retrieved information (bits per pixel). The information in the noisy cue pattern is plotted for comparision (dashed black line). C. Examples of retrieved images for different models. Cue pattern noise is 0.3 (cyan line in panel B). }
    \label{fig:tpam_indexing}
\end{figure}

Here, we explore the ability of TPAM to perform error correction of random indexing patterns within a memory network inspired by the sparse distributed memory (SDM) model \citep{kanerva1988sparse}. The original SDM model consists of two feedforward layers of neurons, an indexing stage with random synaptic weights, mapping data points to sparse binary index vectors, and a heteroassociative memory, mapping index vectors back to data points. Our memory architecture deviates from the original SDM model in three regards. First, it uses complex-valued index patterns. Second, synaptic weights in the indexing stage are learned from the data. Third, it consists of an additional third stage, an error-correction stage using a recurrent TPAM, that sits between the indexing stage and heteroassociative memory (similar as in \citep{Aoki2000}; Fig. \ref{fig:tpam_indexing}A). 

In the following, we denote the data matrix as $\mathbf{P} \in \mathbb{R}^{D \times M}$, where $D$ is the dimensionality of the data and $M$ is the number of data points. As in previous sections, we denote the matrix of index vectors (randomly chosen sparse phasor patterns) by $\mathbf{S} \in \mathbb{C}^{N \times M}$ where $N$ is the number of neurons in the TPAM network.

The indexing stage maps incoming data vectors into index patterns. It is a feedforward network of neurons with the synaptic matrix $\mathbf{W}^I \in \mathbb{C}^{N \times D}$.  
A simple \emph{Hebbian} learning rule for heteroassociation is $\mathbf{\tilde{W}}^I =  \mathbf{S} \mathbf{P}^{\top}$. However, to reduce the level of indexing errors due to inherent data correlations, we use learning that involves the pseudo-inverse of the data, $\mathbf{P}^{+} = \mathbf{V} \mathbf{\Sigma}^{-1} \mathbf{U}^\top$, resulting from the singular value decomposition $\mathbf{P} = \mathbf{U} \mathbf{\Sigma} \mathbf{V}^{\top}$.  Specifically, the synapses in the indexing stage are formed according to:
\begin{equation}
    \mathbf{W}^I =  \mathbf{S} \mathbf{V} \mathbf{\Sigma}^{-1} \mathbf{U}^\top = \mathbf{S} \mathbf{P}^{+}
\label{eqn:encode}
\end{equation}
This linear transform performs \emph{pattern separation} by amplifying the differences between correlated data vectors. It thereby produces decorrelated projections from the data space to the index space. 

The output stage of the network is a heteroassociative memory of the data. This is a feedforward network using simple Hebbian learning for mapping an index pattern back into a data vector. To produce real-valued output patterns, the output neural transfer function projects each component to the real part.

In SDM and heteroassociative memories in general, if the indexing or cue patterns are noisy, the quality of the returned data suffers significantly. To improve the retrieval quality in these cases, we store the indexing patterns $\mathbf{S}$ in TPAM. The TPAM performs error correction on the index patterns produced by the indexing stage before the heteroassociative memory stage.

Empirical comparisons of image storage and retrieval using a simple Hebbian heteroassociative memory, an SDM, and the full network with pattern separation and TPAM for error correction (Fig. \ref{fig:tpam_indexing}B; see Supplement for implementation details) were performed with simulation experiments. The simple Hebbian model and the SDM were also extended by incorporating pattern separation in the indexing stage (solid lines in Fig. \ref{fig:tpam_indexing}B include pattern separation). We stored $M=20$ image patches of $D=12\times12\times3$ pixels into the networks, and measured the correlation $\rho$ of the retrieved pattern with the true pattern given a noisy input cue. We compute the total information per pixel as $I^H = - \frac{1}{2} \log_2 (1 - \rho^2)$ (see Supplement). The full network returns a larger amount of information about the stored data than the simpler models. 
Errors in retrieved TPAM patterns (Fig. \ref{fig:tpam_indexing}C) are due to spurious local minima, which are usually superpositions of stored memories. Similarly, errors in SDM are spurious activations of incorrect patterns, leading to readout errors also as superpositions of stored memories. Including the pseudo-inverse for indexing (PINV), improves the likelihood of avoiding such superposition errors.

\subsection*{Relating TPAM networks to spiking neural networks}
\label{section:TPAMtospikes}
Here, we exploit a natural link between a complex state space and a spike raster through a \emph{phase-to-timing} mapping.
To approximate TPAM networks with networks of integrate-and-fire neurons, we propose biologically plausible mechanisms for the key computations: complex synaptic multiplication, summation of complex postsynaptic signals [\ref{dend_sum}], and the neural transfer function with dynamic threshold [\ref{neural_transfer}].

 \subsubsection*{Phase-to-timing mapping}
\begin{figure}[ht]
\centering
\includegraphics[width=0.45\textwidth]{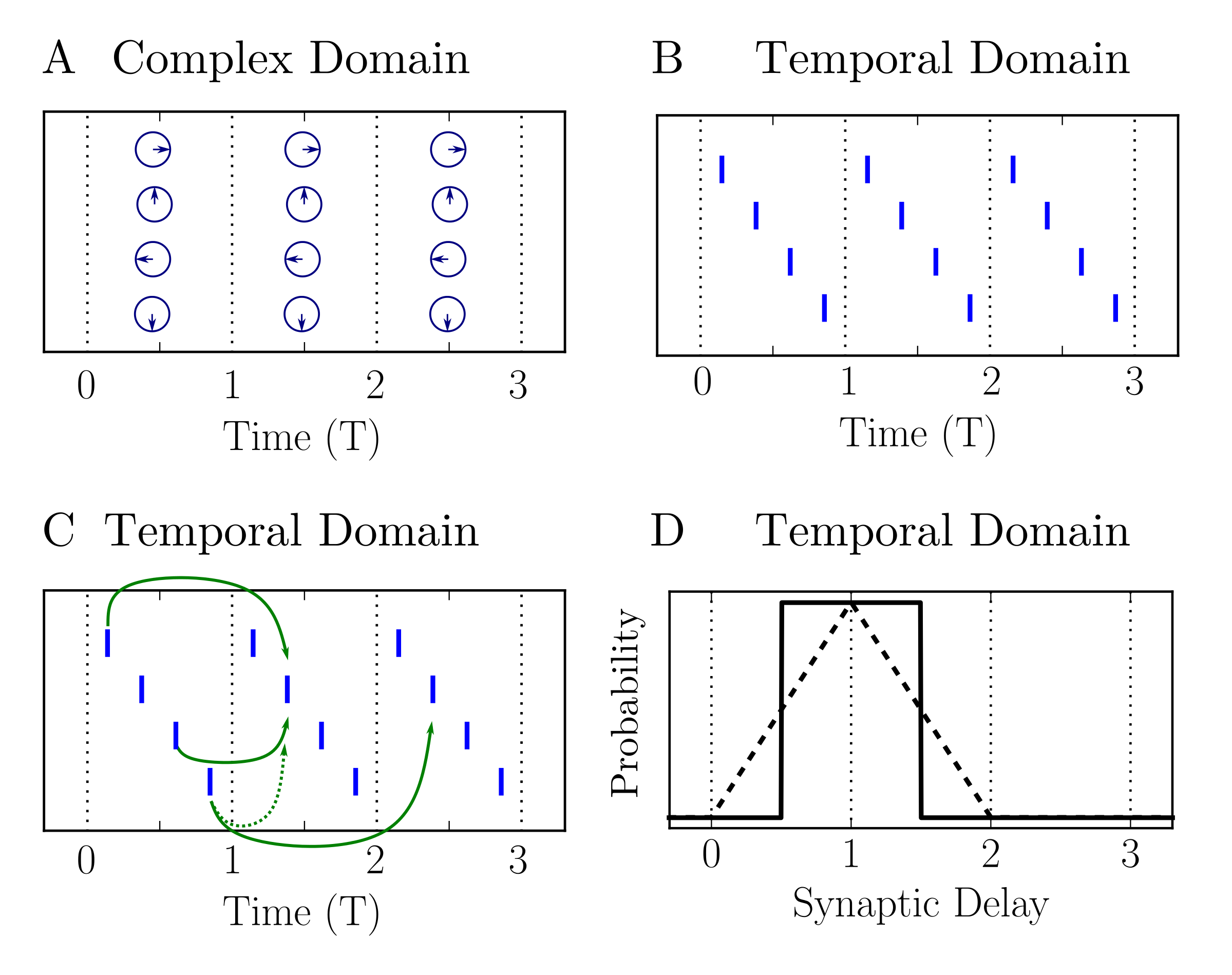}
\caption{\textbf{Mapping states and synapses of TPAM to spiking networks.}
A, B. The complex TPAM state vector (A) can be mapped to a pattern of precisely timed spikes (B) using the phase-to-timing mapping.
C. Influences between spikes after mapping the complex synapses in TPAM to time delays (solid lines). These influences do not obey the discrete time dynamics of TPAM. At any stable fixed point (as shown), a cycle time $T$ can be subtracted or added to individual synapses (dashed lines) so that the discrete time dynamics is obeyed. This manipulation does not change the postsynaptic signals. 
D. The manipulation that matches discrete time dynamics results in a delay distribution between $0$ to $2T$ (dashed).  Because delay synapses can only be adjusted to exactly implement a discrete time dynamics for one particular pattern, we construct delay weights that statistically minimize the violation of the discrete update dynamics across all possible patterns. The delay distribution in our model is (for storing random phases) uniform between $0.5T-1.5T$ (solid).
} 
\label{fig:phase2spikes}
\end{figure}

The complex state vector of TPAM (Fig. \ref{fig:phase2spikes}A) can be uniquely mapped to a spike pattern in a population of neurons (Fig. \ref{fig:phase2spikes}B) through a phase-to-timing mapping. 
The phase angle of a component is represented by the timing of a spike within an interval $T$, where the times between $0$ and $T$ represent the phase angles between $0$ and $2\pi$. A stable fixed point of a complex TPAM state corresponds to a limit-cycle of precisely timed spiking activity, where neurons fire periodically with a rate of $\omega=1/T$ or are silent. The cycle period can be chosen arbitrarily, and is $T=200ms$ in the following simulation experiments.

Note that the phase-to-time mapping is not bijective. To map back from spike rasters to complex vectors, one needs to specify the offset that divides the time axis into intervals of length $T$. Different choices of offsets lead to different sequences of complex vectors. At fixed points, the resulting vector sequences just differ by a global phase shift. Away from fixed points, however, different offsets can also lead to vector sequences with different amplitude structure.

 \subsubsection*{Complex synaptic multiplication}
Via the phase-to-timing mapping, it is also possible to translate the synaptic phasor interactions of a given TPAM to synapses between spiking neurons. Specifically, a phase shift of a complex TPAM synapse translates into a synapse with a specific conduction delay.
The synaptic multiplication then corresponds to the summation of spike-time and synaptic delay.
Connecting neurons with a deterministic spiking mechanism (such as integrate-and-fire) with this synaptic structure results in a network in which neural updates are event-driven and evolve continuously. 
To make all interactions causal, a cycle time $T$ can be added to synaptic delays that would otherwise be negative.

In the TPAM with discrete time dynamics, the new state of a neuron is exclusively a function of the previous time state. 
Thus, for a given phase-to-timing mapping, spike responses in one time window should only depend on spikes within the immediately preceding time window.
For any fixed point, multiples of $T$ can be added or subtracted to the synaptic delays to match the discrete time dynamics (Fig. \ref{fig:phase2spikes}C; dashed). 
However, this multiple of $T$ depends on the particular fixed point, and the mapping to discrete time dynamics cannot be simultaneously guaranteed for multiple fixed points stored in the network. 

Here, we selectively increment delays by $T$ in order to wrap the synaptic delay values to a range between $0.5T$ and $1.5T$. This choice maximizes interactions between subsequent time steps, but it cannot fully eliminate interactions within the same timestep or across two timesteps (Fig. \ref{fig:phase2spikes}C, D; solid). 
Yet, at any fixed point this choice of synaptic delays produces exactly the same postsynaptic potentials as the synaptic delays that correspond to discrete time dynamics. 
Thus, even though the spiking network does not strictly obey a discrete time dynamics, at each equilibrium state an equivalent network exists that satisfies discrete time dynamics and produces exactly the same postsynaptic inputs to all neurons.

\subsubsection*{Postsynaptic complex vector summation}
The complex sum of TPAM [\ref{dend_sum}] corresponds to the addition of sine waves with different phases and amplitudes that oscillate with period $T$ in the time domain. 
To perform this computation with a spiking neural network, the effect of each spike should produce a phase-locked oscillatory zero-mean postsynaptic current. 
In our model of standard integrate-and-fire neurons \citep{Goodman2009, Eliasmith2003, Izhikevich2007},  
the postsynaptic oscillation caused by each spike is generated by a monosynaptic excitatory current, as well as inhibitory current routed through interneurons. 
The action-potential and the opening of synaptic channels act on a fast time-scale, which allows these kinetic features to be simplified as instant jumps in the dynamic state variables.
When a neuron's membrane potential reaches threshold $V_\theta$, it fires a spike and the membrane potential is set to the reset potential $V_r$. After the synaptic delay, the spike causes postsynaptic channels to open. This is modeled as a jump in a synaptic state variable $s_{ij}$ that injects current proportional to the synaptic magnitude $I_{i} = \sum_j |W_{ij}| s_{ij}$, which then decays exponentially as the channels close (see Supplement for details).

\begin{figure}[ht]
\centering
\includegraphics[width=0.65\textwidth]{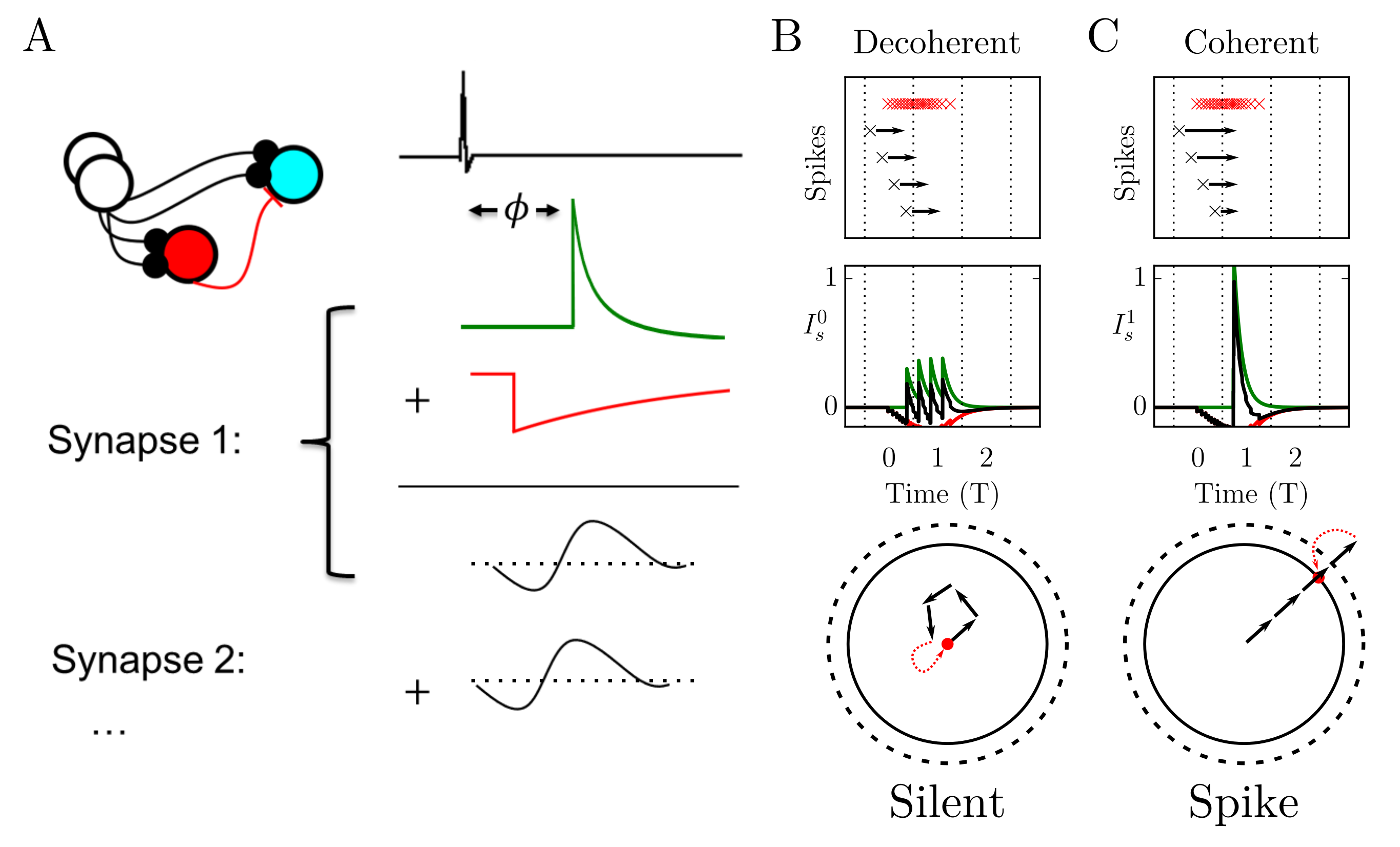}
\caption{\textbf{Mechanisms to implement TPAM with integrate-and-fire neurons.} A. Direct excitation and indirect inhibition produces a postsynaptic current oscillation. B, C. The effect of a presynaptic spike raster (Row 1; red x's: inhibitory neuron spikes) depends on the phase relationships of individual postsynaptic currents. The resulting postsynaptic currents (Row 2) can be rather small (decoherent, B), or quite large (coherent, C) (green: excitation, red: inhibition, black: total). If the current is large enough, the neuron will reach threshold and fire a spike.}
\label{fig:spike_mech}
\end{figure}

The time constants determining the decays of neural and synaptic variables are tuned proportional to the cycle time $T$ to create the oscillatory current. The time constant of the membranes of excitatory neurons is $C/g_l = 0.25T$, and $0.1T$ for interneurons. 
The time constant for inhibitory and excitatory synapses is $T$ and $0.5T$, respectively. 
With these settings, the total postsynaptic current elicited by a spike
forms (approximately) a sine wave oscillation with cycle period of $T$ (Fig. \ref{fig:spike_mech}A).

The excitatory synapses have individual synaptic delays determined by the phases of the recurrent weight matrix $\mathbf{W}$, whereas the delays to inhibitory neurons are all the same. Because inhibitory neurons operate linearly and their synaptic delays are nonspecific, a single inhibitory neuron (or arbitrary sized population) can serve to route inhibition for multiple complex synaptic connections.
The delays of the excitatory synapses determine the phase of the oscillatory currents. Depending on presynaptic spike times and synaptic delays, the currents can sum either decoherently or coherently (Fig. \ref{fig:spike_mech}B, C), estimating the complex sum. Altogether, the complex dot product is approximated by the spiking network.

\subsubsection*{Neural transfer function and threshold control}
The inhibitory neurons serve a second purpose besides shaping the postsynaptic oscillatory currents. They also account for the global normalization needed for the threshold strategy [\ref{neural_transfer}], which keeps the activity sparse. The inhibitory population integrates the activity from the excitatory neurons and routes inhibition back into the population in proportion to the overall activity.
The magnitude of this feedback inhibition can be tuned to limit the activity of the population, and it implements the threshold strategy needed for TPAM, e.g. $\theta |\mathbf{z}|$ [\ref{eqn:threshold_strat}]. 
The gain of the inhibitory feedback is modulated by several mechanisms: the gain of the E-to-I synapses, the gain of the I-to-E synapses, the time constants of the synapses, the number of inhibitory neurons, and the gain of the inhibitory neural transfer function. Each of these gain factors can be analytically accounted, with a linear approximation being useful to understand the gain of the spiking neural transfer function (see Supplement).

The deterministic dynamics of the integrate-and-fire neuron will cause the neuron to spike whenever the current drives the voltage above threshold. If the magnitude of the input current oscillation is large enough, then the neuron will fire at a consistent phase.
For the excitatory neurons, a refractory period slightly over half the cycle time (i.e. $\tau_{ref} = 0.5T$) acts as the Heaviside function on the magnitude. This implements the phasor-projection of TPAM by limiting the neuron to one spike per cycle, while preserving the phase through the spike timing [\ref{neural_transfer}]. The parameter value $V_{\theta}$ sets the threshold $\theta$ of whether the neuron will fire or not [\ref{eqn:threshold_strat}].

\subsubsection*{Simulation experiments with spiking networks}

\begin{figure}[t]
    \centering
    \includegraphics[width=0.9\textwidth]{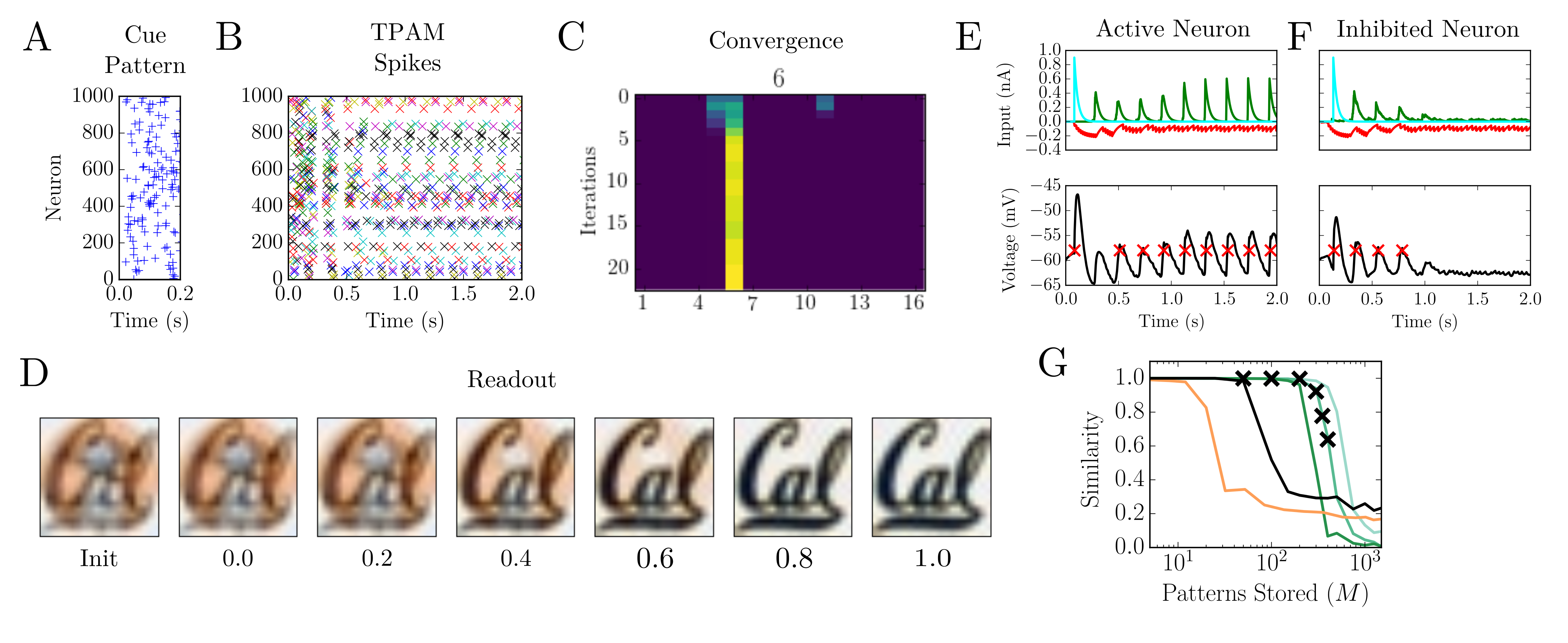}    \caption{
    \textbf{Memory network with sparse spike timing patterns.} The network is cued with a spiking pattern that encodes a superposition of three stored images (A). B. After a short time the network converges to a stored attractor state, a sparse spiking pattern that persistently repeats. C. Evolution of similarities between network state and index patterns of different images,
    a high match is indicated by yellow. One pattern out-competes the other two after a few iterations. 
    D. Retrieved image as a function of time.
    E.  Postsynaptic currents in a neuron that is active in the attractor state. The neuron receives initialization input current (top; cyan), and participates in the stable oscillation. The excitatory current (green) is balanced by the inhibitory current (red). The cell takes part of the coherent phase pattern and begins oscillating and spiking (bottom; red x's) at a consistent phase relative to the rest of the population. F. Postsynaptic currents in a neuron that receives the initialization input, but is quiescent during the stable oscillation. The inhibitory current outweighs the decoherent excitatory current and the neuron is silent. G. Retrieval performance of the spiking implementation of TPAM, measured by average similarity between retrieved and target state as function of stored patterns  (black x's). The performance matches the performance of a TPAM  with a fixed threshold (green lines). The spiking network can store more patterns than the traditional Hopfield network (black) \citep{Hopfield1982} or the dense phasor network (orange) \citep{noest1988phasor}.
    }
    \label{fig:spike_tpam}
\end{figure}

For storing a small collection of RGB images, a network architecture with spiking neurons was implemented as depicted in Fig. \ref{fig:tpam_indexing}A. The indexing stage transforms the real-valued input image into a complex vector from the encoding matrix (see Supplement). The complex vector is mapped into a timed sequence of input spikes (Fig. \ref{fig:spike_tpam}A), which is the initial input to a spiking TPAM network. The spiking TPAM network is decoded with a Hebbian heteroassociative memory. This output stage uses the same synaptic mechanisms to implement the complex dot product as described for TPAM. However, the readout neurons respond proportionally to the input magnitude (i.e. no refractory period, see Supplement). 

In the simulation experiment, the network is cued with several overlapping patterns and noise. After initialization through the indexing stage, the spiking activity in the TPAM network quickly settles into an oscillatory pattern (Fig. \ref{fig:spike_tpam}B), which corresponds to the index of one of the stored patterns (Fig. \ref{fig:spike_tpam}C). 
Interestingly, the global oscillation of the network is generated intrinsically -- it does not require any external driving mechanism. 
The output of the heteroassocitive memory stage during the convergence of the TPAM dynamics (Fig. \ref{fig:spike_tpam}D) shows how the dynamics rapidly settles to one of the stored patterns superposed in the input, out-competing the other two.

Examples of the dynamics of individual neurons in the network are illustrated. 
A neuron participating in the stable oscillation receives coherent volleys of excitatory input and has a large oscillation in its membrane potential (Fig. \ref{fig:spike_tpam}E). Another neuron shown is initially excited, but then becomes silent in the stable attractor state (Fig. \ref{fig:spike_tpam}F). In this neuron excitation becomes decoherent, and as a consequence the voltage is never large enough to overcome threshold.

Finally, the capacity of the spiking network is examined in simulation experiments (Fig. \ref{fig:spike_tpam}G). The spiking network is robust without much parameter tuning and not perfectly optimized, but still retrieval performance based on the number of stored patterns (Fig. \ref{fig:spike_tpam}G, black x's) easily exceeds the performance of a traditional bipolar Hopfield network (Fig. \ref{fig:spike_tpam}G, black line). The performance curve of the spiking model is nicely described by the performance curve of the complex TPAM with similar settings. However, the spiking network does not reach the full performance of the complex TPAM with optimized threshold. The sparsity and threshold parameters of the spiking network were not fully optimized in our experiments. But also, some of the approximations in the spiking network add noise, which prevents it from reaching the full capacity of the ideal complex model.  Nonetheless, these experiments show that the spiking model does behave in a manner that can be captured by the complex TPAM model. 

\subsection*{Sequence associative memories and complex attractor networks}
Last, we investigate how complex fixed point attractor networks can help to understand sequence associative memories: simple networks with binary threshold neurons and parallel, time-discrete update dynamics for storing sequences of patterns of fixed length (see Background). Consider the storage of a closed sequences or limit cycles of fixed length $L$: $\mathbf{\xi}^1 \to \mathbf{\xi}^2  \to ... \to \mathbf{\xi}^L \to \mathbf{\xi}^1 \to ...$, with $\mathbf{\xi}^i \in \mathbb{R}^N \;\forall i = 1, ..., L$. In case of storing multiple sequences an index is added to label the different sequences: $\{\xi^{\mu,s}, \mu=1,...,M\}$.   
The learning in these models is also described by a Hebbian outer-product learning scheme \citep{Amari1972}. Here we use a combination of Hebbian and anti-Hebbian learning to produce a skew-symmetric interaction matrix:
\begin{equation}
\mathbf{J} = \sum_{\mu=1}^M \sum_{s=1}^L \mathbf{\xi}^{\mu, s} \left(\mathbf{\xi}^{(\mu, s - 1) \bmod L}- \mathbf{\xi}^{(\mu, s + 1) \bmod L}\right)^{\top}  
\label{amari_sequence_basic}
\end{equation}

Since the matrix $\mathbf{J}$ is skew-symmetric, there is no Lyapunov function describing the limit cycle dynamics in the network. However, we can use the spectral properties of the weights to construct an equivalent fixed point attractor network. 

Consider [\ref{amari_sequence_basic}] for the simple example with $L=N=3$, $M=1$ and the stored patterns $\xi$ being the cardinal basis vectors of $\mathbb{R}^3$. One of the complex eigenvectors of $\mathbf{J}$ is $\mathbf{v}= (e^{\textrm{i} \frac{2\pi}{3}}, e^{\textrm{i} \frac{4\pi}{3}}, 1)^{\top}=:(e^{\textrm{i} \phi_1}, e^{\textrm{i} \phi_2}, e^{\textrm{i} \phi_3})^{\top}$, which is the (equidistant) phasor pattern that represents the entire stored limit cycle in complex space. 
One can now form a complex matrix $\mathbf{W'}$ that posesses $\mathbf{v}$ as a fixed point, i.e., has eigenvalue of one, simply by dividing $\mathbf{J}$ by the eigenvalue associated with $\mathbf{v}$, which is $\lambda = \textrm{i} \sqrt{3}$:   
\begin{equation}
\mathbf{W'}=\frac{1}{\textrm{i} \sqrt{3}} \mathbf{J} = \frac{1}{\textrm{i} \sqrt{3}}
\begin{bmatrix}
    0      & 1 & -1\\
    -1   & 0 & 1\\
    1   & -1 & 0
\end{bmatrix}
\label{sequence_phasor}
\end{equation}
Since the eigenvalues of any skew-symmetric matrix have zero real part \citep{eves1980elementary} the interaction matrix $\mathbf{W'}$ is always Hermitian in general. Thus, the described construction is a recipe to translate sequence memory networks into complex neural networks governed by a Lyapunov dynamics. In the resulting networks, the synaptic matrix is $\mathbf{W'}$, the neural nonlinearity is $g(u_i) = u_i/|u_i|$, and the Lyapunov function is [\ref{phasor_lyapunov}].  

One could now suspect that storing the pattern $\mathbf{v}$ in a phasor network \citep{noest1988phasor} would result in the same network interaction matrix $\mathbf{W'}$. However, this is not the case. The weight matrix resulting from learning the phase vector $\mathbf{v}$ with the conjugate outer-product learning rule [\ref{phasor_learning}] is:
\begin{equation}
\mathbf{W} = \mathbf{v} \mathbf{v}^{*\top} - \mathbb{1} =
\begin{bmatrix}
    0      & e^{\textrm{i}(\phi_1-\phi_2)} & e^{\textrm{i}(\phi_1-\phi_3)}\\
    e^{\textrm{i}(\phi_2-\phi_1)}   & 0 & e^{\textrm{i}(\phi_2-\phi_3)}\\
    e^{\textrm{i}(\phi_3-\phi_1)}   & e^{\textrm{i}(\phi_3-\phi_2)} & 0
\end{bmatrix}
\label{w_example}
\end{equation}
The phase vector $\mathbf{v}$ is again an eigenvector of $\mathbf{W}$. 

The take away from this example are the following points:

\noindent
\textbf{\textit{(1.)}} Outer-product sequence memories with skew-symmetric weights can be mapped to continuous-valued phasor networks with Hermitian weights $\mathbf{W'}$, whose dynamics is described by the Lyapunov function [\ref{phasor_lyapunov}]. 
A similar idea of deriving a Lyapunov function for $M$-periodic sequences of binary patterns was proposed in \citep{herz1991global, herzetal1991statistical}. Specifically, these authors proposed to embed sequence trajectories in a real-valued space, consisting of $M$ copies of the original state space.
    
\noindent
\textbf{\textit{(2.)}} Continuous-valued phasor networks derived from sequence memories with outer-product rule [\ref{amari_sequence_basic}] {\it are different} from phasor networks using the conjugate complex outer-product learning rule [\ref{phasor_learning}], such as TPAM networks.    
    
\noindent
\textbf{\textit{(3.)}} Phasor networks derived from outer-product sequence memories have two severe restrictions. First, since the synaptic weights are imaginary without real part, they only can rotate the presynaptic signals by 90 degrees. Second, the first-order Markov property of sequence memory networks translates into phasor networks with interactions only between direct phase angle neighbors. 

\noindent
\textbf{\textit{(4.)}} Phasor networks derived from sequence memories can, by construction, only store phase patterns whose phase angles are equidistantly dividing $2\pi$. Because of symmetry reasons, such equidistant phase patterns can be stabilized in spite of the restrictions described in the previous point.

\section*{Discussion}
\label{sec:discussion}

We present a theory framework for temporal coding with spiking neurons that is based on fixed point attractor dynamics in complex state space. Our results describe a new type of complex attractor neural network, and how it can be used to design robust computations in networks of spiking neurons with delay synapses. 

\subsection*{Threshold phasor networks}
A novel type of complex attractor network is introduced, threshold phasor associative memory (TPAM). TPAM inherits from previous complex fixed point attractor networks the Lyapunov dynamics,
and the capability to store arbitrary continuous-valued phasor patterns. 
The neural threshold mechanism added in TPAM yields advantages over existing phasor memory networks \citep{noest1988phasor}. 
First, it significantly increases the memory capacity, the amount of information that can be stored per synapse, as we show in simulation experiments.
Second, the amplitude threshold operation in complex space prevents neurons to represent phase angles whose postsynaptic phase concentration is diffuse. Therefore, all phase angles in a TPAM state vector carry high information about the postsynaptic inputs. 
Third, TPAM networks describe physiologically realistic sparse firing patterns. 

The ability to store continuous phasor patterns as attractor states permits the processing of analog data in TPAM without preceding data discretization. 
However, data correlations still causes problems in TPAM networks as in other associative memory models. 
For the storage of correlated sensor data, such as images, we propose a three-layer network extending from previous memory indexing models, such as sparse distributed memory \citep{kanerva1988sparse}.
We demonstrate in an image retrieval task that a network consisting of a pattern separation stage, a TPAM network for error correction, and a heteroassociative memory has improved performance, compared to earlier models.

\subsection*{Mapping complex attractors into periodic temporal codes}
Previous models of spike-timing based computation often proved brittle, which even served as an argument for rendering spike-timing codes as irrelevant for neuroscience \citep{london2010sensitivity}.
We employ a straight-forward \emph{phase-to-timing} mapping to translate the complex fixed point attractors in TPAM to periodic patterns of precisely timed spiking activity. 
Although the time-continuous nature of spike interactions through delay synapses and the time-discrete dynamics in TPAM are not equivalent in the entire state space, they are so at the fixed points. 
The correspondence between complex fixed point states and periodic spiking patterns
enables robust computation with spike-timing codes, as we demonstrate with integrate-and-fire neural networks.
Such periodic spiking patterns are (superficially) similar to and compatible with spike-timing based codes proposed previously. For example, Hopfield \citep{Hopfield1995} proposed a network for transducing rate-coded signals into periodic spike-timing patterns, which encodes the real-valued input into the phase of a complex number. 
Other examples are \emph{synfire braids} \citep{Bienenstock1995} and \emph{polychronization} \citep{Izhikevich2006} in networks with synaptic delays and neurons that detect synchrony. 

While TPAM provides valuable insights into network computations using the timing of single spikes, a simple extension is able to also capture aspects of rate coding.
By relaxing the constraint on binary magnitudes in TPAM,  complex values with variable magnitudes can be represented, which correspond to variable firing rates.
For instance, if the neural response magnitude is a saturating sigmoid function, attractors would lie near saturation points, as in the real-valued Hopfield model \citep{Hopfield1984}. Yet, phase values can still be arbitrary. 
The corresponding spike patterns would still be oscillatory, with bursts at a particular phase followed by durations of silence in the anti-phase. This type of burst spiking is indeed seen in hippocampal place cells that are locked to the theta rhythm \citep{OKeefe1993}.

Further, it is shown how Hebbian sequence memories in discrete time \citep{Amari1972} can be described by fixed point dynamics in the complex domain.
This description is similar to the construction of a Lyapunov function for sequence memories in an enhanced (real-valued) state space \citep{herzetal1991statistical}. 
Interestingly, the complex attractor networks corresponding to sequence memories [\ref{sequence_phasor}] are different from phasor networks with conjugate outer-product Hebbian learning [\ref{phasor_learning}].

\subsection*{Complex algebra with spiking neural circuits}
We describe how the postsynaptic computation in TPAM neurons can be implemented by circuits of excitatory and inhibitory leaky integrate-and-fire neurons. In essence, the required complex dot product can be achieved by shaping the postsynaptic currents caused by individual spikes into (approximately) a sinusoid. 
The shaping is achieved through a combination of dendritic filtering, synaptic delays and adding inhibitory neurons that balance excitation. 
The required mechanisms are in accordance with the following observations in neuroscience: 
\emph{Periodic firing and synchrony:} Action-potentials in the brain are often periodic, synchronized with intrinsic rhythms visible in local field potentials
\citep{Buzsaki2004,Riehle1997,Lynch2016}.
\emph{Delayed synaptic transmission:} Due to variability in axon length and myelination, the distribution in measured delay times in monosynaptic transmission is broad, easily spanning the full cycle length of gamma and even theta oscillations \citep{Swadlow2000}. 
\emph{Balance between excitation and inhibition:}
Excitation/inhibition balance is widely seen throughout cortex \citep{Marino2005, Haider2006, Atallah2009}, and inhibitory feedback onto pyramidal cells is a major feature of the canonical cortical microcircuit \citep{Douglas1989, Isaacson2011}.

Complex algebra with spiking neurons is relevant for other types of complex neural networks as well, such as working memory networks based on the principles of reservoir computing \citep{Frady2018}. Because spiking neurons represent both the real and imaginary parts, complex spiking reservoir networks would have twice the memory capacity per neuron than rate-based models. Further, TPAM-like networks with modified learning rules could potentially be used to construct spike-timing implementations of line-attractors--useful for understanding place coding in hippocampus \citep{Welinder2008, Campbell2018}, or for modeling/predicting dynamical systems \citep{Eliasmith2012, Deneve2017}. 

The conjugate outer-product learning rule in TPAM requires tunable synaptic delays, as suggested in previous models \citep{huningglunder1998synaptic}, but without compelling biological evidence.
Alternatively, the wide distribution of fixed synaptic delays in brain tissue could enable biologically observed spike-timing-dependent plasticity (STDP) \citep{Szatmary2010} to shape synaptic connectivity compatible with the conjugate outer-product learning rule.
An interesting subject of future research is the effect of STDP mechanisms in the presence of oscillations and synaptic delays, and how it relates to outer-product Hebbian learning rules.

The presented theory also has impact on \emph{neuromorphic computing}. Several groups have been exploring and engineering spiking neural networks as a new paradigm for computation, such as \emph{Braindrop} \citep{Neckar2019} 
and IBM's \emph{True North} \citep{Merolla2014}. Recently, Intel announced the neuromorphic chip \emph{Loihi} \citep{Davies2018}, with features such as individual synaptic delays and on-chip learning. Our theory offers a principled way of ``programming'' spiking-neuron hardware, leveraging the speed of temporal codes and providing straight-forward connections to complex matrix algebra and Lyapunov dynamics.

\section*{Acknowledgements}
This work was supported by the Intel Corporation (ISRA on Neuromorphic
Architectures for Mainstream Computing), NSF award 1718991, and Berkeley DeepDrive. The authors would like to thank Pentti Kanerva and members of the Redwood Center for valuable feedback.

\section*{Supplemental Methods}

\subsection*{Threshold phasor associative memories}
\subsubsection*{Derivation of the Lyapunov function for TPAM}
To derive a Lyapunov function for TPAM, we first consider Hopfield networks with real-valued neurons. Hopfield \citep{Hopfield1984} showed that for an invertible neural transfer function $f(x)$, the corresponding Lyapunov function is:
\begin{equation}
E(\mathbf{x}) = - \frac{1}{2} \sum_{ij} J_{ij} x_i x_j + \sum_i \int_0^{x_i} f^{-1}(v) dv
\label{cont_hopfield}
\end{equation}
The first term is the energy function of the Hopfield network with binary neurons \citep{Hopfield1982}.  
The derivative of the second term in [\ref{cont_hopfield}] is just $f^{-1}(x_i)$. Thus, setting the derivative with respect to $x_i$ to zero leads to coordinate-wise update equations with transfer function $f(x)$: $x_i(t+1) = f(\sum_j J_{ij} x_j(t))$. 

To understand the behavior of binary neurons with non-zero thresholds, we consider the neural transfer function $H(x_i-\Theta_i)$, where $\Theta_i$ is the constant, individual threshold of neuron $i$. Although the transfer function is not invertible, it can be approximated by an invertible function, for example, the logistic function. In the limit of making the approximation tight, i.e., $f(x)\approx H(x_i-\Theta_i)$, the Lyapunov function [\ref{cont_hopfield}] becomes
\begin{equation}
E(\mathbf{x}) = - \frac{1}{2}\sum_{ij} J_{ij} x_i x_j + \Theta^T \mathbf{x} +b(\mathbf{x}) 
\label{lyapunov_hopfield_indi_thres}
\end{equation}
Thus, the second term in [\ref{cont_hopfield}] decomposes into a barrier and a bias term.
The potential barrier term is $b(\mathbf{x})=\infty \; (1- \prod_i H(1-x_i) H(x_i) )$, which blocks components to assume any values outside the interval $(0,1)$. The other is a bias term, well known from the Ising model. Note that both additional terms in [\ref{lyapunov_hopfield_indi_thres}] vanish for governing the updates of binary threshold neurons with $\Theta_i=0 \; \forall i$, leading to the energy function of the orginial Hopfield model \citep{Hopfield1982}. The barrier term is unnecessary because enforced implicitly by any binary-valued neural update.

If all neurons in the network experience the same global threshold, $\Theta = \Theta_i \;\forall i$, and the barrier term enforces all state components to lie within the $(0,1)$-interval, the bias term becomes an $L_1$ constraint: 
\begin{equation}
E(\mathbf{x}) = - \frac{1}{2}\sum_{ij} J_{ij} x_i x_j + \Theta ||\mathbf{x}||_1 +b(\mathbf{x}) 
\label{lyapunov_hopfield_thres}
\end{equation}
For $\Theta>0$ the $L_1$ term penalizes states with non-zero components, encouraging sparser states with fewer active neurons.    

Further, if the threshold control is dynamic, with the threshold a linear function of the network activity, $\Theta = \theta \sum_i x_i$, the Lyapunov function becomes:
\begin{equation}
E(\mathbf{x}) = \sum_{ij} \left( - \frac{1}{2} J_{ij} + \theta \mathbb{I} \right) x_i x_j  +b(\mathbf{x}) \label{lyapunov_hopfield_dyn_thres}
\end{equation}
Thus, a linear dynamic threshold control corresponds to a antiferromagnetic term added to the network interactions, which can be easily modeled by inhibition. 

The Lyapunov functions derived above can be easily be generalized to TPAM. 
In analogy to [\ref{cont_hopfield}], the energy function of the phasor neural network \citep{noest1988phasor} for arbitrary invertible transfer function $f(z)$ can be extended to: 
\begin{equation}
E(\mathbf{z}) = - \frac{1}{2}\sum_{ij} W_{ij} z_i z_j^* + \sum_i \int_0^{|z_i|} f^{-1}(v) dv
\label{lyapunov_ext}
\end{equation}
The neural transfer function of TPAM, $g(z;\Theta)$ can be approximated by an invertible function $f(z)$ by replacing the Heaviside function in the neural transfer function with an invertible function, such as the logistic function. 
In the limit of making the approximation tight, i.e., $f(z)\approx g(z;\Theta)$,
and with constant threshold $\Theta = \Theta(t)$ the Lyapunov function [\ref{lyapunov}] becomes
\begin{equation}
E(\mathbf{z}) = - \frac{1}{2}\sum_{ij} W_{ij} z_i z_j^* + \Theta|\mathbf{z}| +b(\mathbf{z}) 
\label{lyapunov_tpan_ext}
\end{equation}
the original energy function of phasor neural networks \citep{noest1988phasor} with, again, two additional terms. The barrier function in this case is $b(\mathbf{z})=\infty \; (1- \prod_i H(1-|z_i|))$, the bias term is again a $L_1$ term, which encourages networks with lower activity. 

In analogy with [\ref{lyapunov_hopfield_dyn_thres}], for linear dynamic threshold control in the threshold strategy, the Lyapunov function of TPAM becomes:
\begin{equation}
E(\mathbf{z}) = \sum_{ij} \left(- \frac{1}{2} W_{ij} + \theta \mathbb{I} \right) z_i z_j^* +  +b(\mathbf{z})
\label{lyapunov_tpan_dyn_ext}
\end{equation}
Thus, the transition from a constant threshold setting to a dynamic threshold control replaces the $L_1$ constraint by a self interaction term.

\subsubsection*{Comparison of TPAM to other associative memory models}
\begin{figure}
    \centering
    \includegraphics[width=0.5\textwidth]{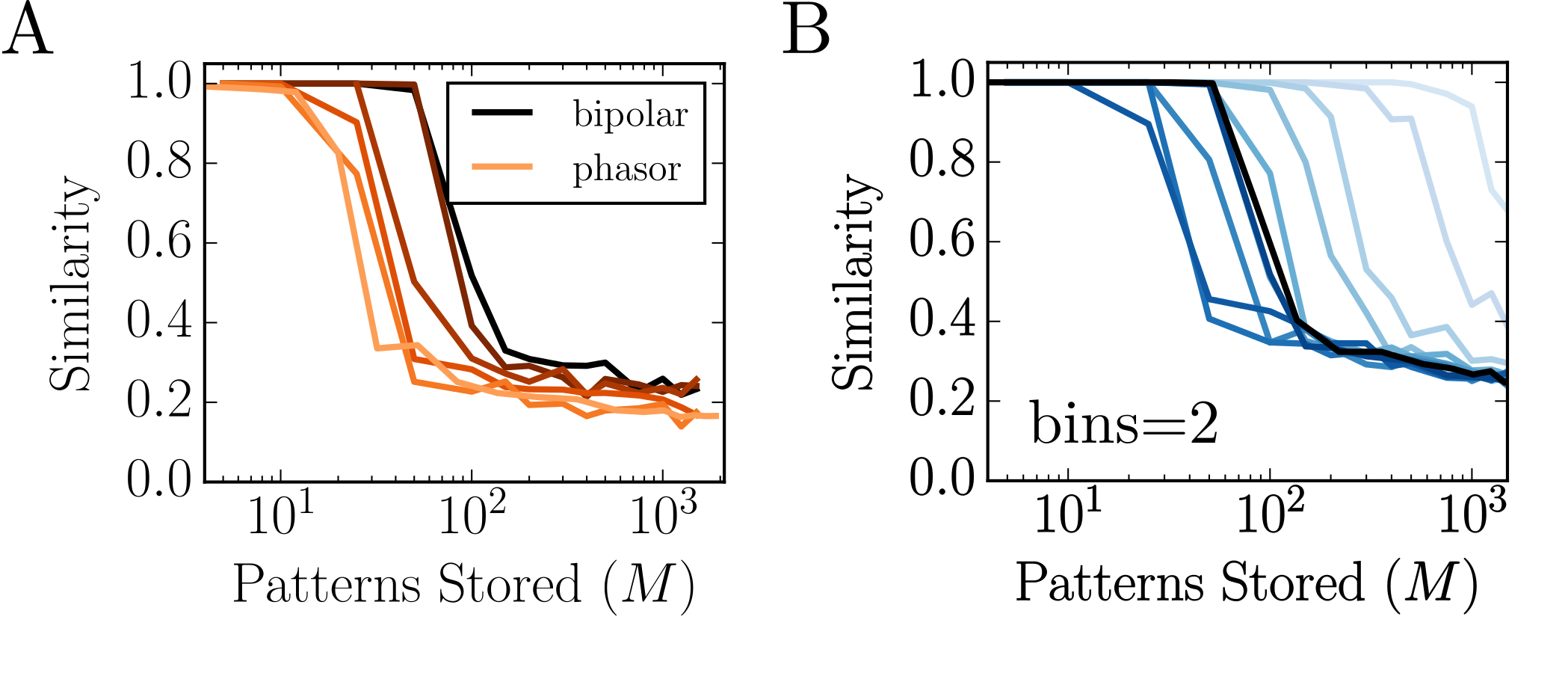}
    \caption{\textbf{Capacity of other associative memory networks.} A. Capacity of discrete phasor networks (csign). Black is two bins (phase 0 and phase pi) equivalent to a bipolar Hopfield network. Orange is continuous phasor network with infinity bins. Intermediate colors are increasing bin sizes $L = [2,4,6,8,12]$.
    B. Capacity of ternary sparse memory networks. Black from A. Blue lines are memory models with lighter colors indicating increasing sparsity.
}
    \label{fig:ann_capacity}
\end{figure}

We performed capacity experiments for previous models of associative memory to compare with TPAM.
In \ref{fig:ann_capacity}A, we compare
a phasor memory network with continuous phase variables (without threshold) to complex Hopfield networks with discretized phase representations, in which the full phase circle is equidistantly divided into $L$ bins \citep{Aoki2000}. For $L=2$, this becomes the traditional bipolar Hopfield network. 
For all models the similarity is high at small $M$, then falls off rapidly. In models with larger numbers of bins, the drop-off of the similarity starts at smaller numbers of stored patterns. 

Fig. \ref{fig:ann_capacity}B compares threshold phasor networks with binary phase discretization. The black line is the Hopfield model without threshold storing dense bipolar patterns (same as the black line in subfigure A). The blue lines correspond to models with threshold storing sparse ternary patterns containing $-1, 0, 1$ components (the lighter the color, the sparser the patterns). 
For ternary attractor networks, the capacity decreases with moderate sparsity values (dark blue lines below the black line). As the sparsity increases further, the performance supersedes the bipolar Hopfield network (light blue lines). Thus, at sufficient levels of sparsity, the thresholded models can store significantly more patterns than the standard Hopfield model. 

\subsubsection*{Information theory for TPAM capacity}
\label{sec:tpam_info_methods}

 To measure the total information in sparse phasor vectors, we treat the information in phases and amplitudes separately. 

The amplitude structure is binary and one has just to estimate the two error types, false positives $\alpha$ and misses $\beta$ in the simulation experiments.  The information needed to correct the errors of a sparse binary vector is given by:
\begin{equation}
    I_{corr} = \hat{p} I \left( \frac{\alpha (1-p_{hot})}{\hat{p}} \right) + (1 - \hat{p}) I \left( \frac{\beta p_{hot}}{1-\hat{p}} \right)
\end{equation}
where $I(p) = - p \log p - (1-p) \log (1-p) $ is the Shannon information, and $\hat{p} = \alpha (1 - p_{hot}) + (1 - \beta) p_{hot}$.

The phase information is estimated from the statistics of phasor vectors observed in the simulation experiments.
A von Mises distribution is fit to the difference between the true phasor and retrieved phasor variable, yielding the phase concentration parameter $\kappa$ which is inversely proportional to the variance. In the high-fidelity regime, when the network has only a few patterns stored, the von Mises concentration parameter $\kappa$ can be approximated with a Gaussian fit. We measured $\kappa$ empirically from simulation experiments. The entropy of the von Mises distribution based on $\kappa$ is:
\begin{equation}
    VM(\kappa) = \log( 2 \pi I_0 (\kappa) ) - \kappa \frac{I_1(\kappa)}{I_0(\kappa)}
\end{equation}
and the information of a phasor variable is:
\begin{equation}
    I_{phase} = \log 2 \pi - VM(\kappa)
\end{equation}

The total information for a single item stored in memory is then the information for each phasor variable plus the information of the sparse binary vector:
\begin{equation}
I_{item} =  N \left( I(p_{hot}) - I_{corr} + p_{hot}(1-\beta)I_{phase}  \right)
\end{equation}

Given that the vectors are i.i.d., each item stored yields the same information. The total information is then given by the information per item and the total number of items stored, normalized by the number of synapses:
\begin{equation}
I_{total} = M I_{item} / N^2
\end{equation}

\subsubsection*{Hetero-associative capacity experiments}
\label{sec:hetero_methods}

We compared three models of hetero-associative memories: a simple model with Hebbian learning, the Sparse Distributed Memory, and a novel indexing method with TPAM as an autoassociative clean-up memory. We compared designs that have random indexing to designs that also include pattern separation in the indexing stage (dashed lines versus solid lines, respectively, in Fig. 4B). 

For random indexing, the encoding matrix $\mathbf{\tilde{W}}^I$ is random. For the Hebbian learning model and SDM, the encoding matrix was a random sparse binary matrix. For TPAM the encoding matrix was sparse random phasors with $p_{hot}=10\%$. 
The retrieval of a data vector with a given phasor pattern is $u_i = \sum_j W^H_{ij} z_j$, with $\mathbf{W}^{H}$ the decoding matrix.

The decoding matrix for each method is computed from the activity of the hidden layer. For the Hebbian learning model, $\mathbf{\tilde{H}}_{\mu} ^{(Hebbian)} = \mathbf{\tilde{W}}^I \mathbf{P}_\mu$, where $\mathbf{P}_\mu$ is the cue pattern. The decoding matrix is then $\mathbf{W}^H = \mathbf{P} \mathbf{\tilde{H}}^{(Hebbian) \top}$.
The SDM has a non-linear threshold function in the hidden layer that can provide some clean-up, $\mathbf{\tilde{H}}_\mu^{(SDM)} = g(\mathbf{\tilde{W}}^I \mathbf{P}_\mu)$. In this case, the function $g$ takes the $K$ largest values of the input vector and sets them to 1, with all others set to 0. The decoding matrix is similarly $\mathbf{W}^{H} = \mathbf{P} \mathbf{\tilde{H}}^{(SDM)\top}$. 

For the TPAM with random indexing, a codebook of random phasors is chosen $\mathbf{S} \in \mathbb{C}^{N \times M}$. The encoding matrix  is $\mathbf{\tilde{W}}^I = \mathbf{S}\mathbf{P}^\top$. The index patterns are stored in the auto-associative recurrent matrix $\mathbf{W} = \mathbf{S} \mathbf{S}^{*\top}$. 
The index vector can be decoded with the simple heteroassociative conjugate outer-product learning rule $\mathbf{W}^H = \frac{1}{K} \mathbf{P} \mathbf{S}^{*\top}$

To include pattern separation, the indexing matrices are altered. Rather than being purely random, they include both a random part (the random index pattern) and the pseudo-inverse of the patterns, $\mathbf{W}^I = \mathbf{S} \mathbf{P}^{+}$. We also replaced the original SDM encoding matrix with a pattern separation matrix to understand the effects of pattern separation and clean-up on memory retrieval performance. For Hebbian learning and SDM models, the matrix $\mathbf{S}$ is a random sparse binary matrix. 

The index patterns are random, but in high-dimensions this means that they are approximately orthogonal. 
We thus do not need to worry too much about correlations in the index vectors (however, performance can be increased by ensuring that the index patterns are exactly orthogonal).

If the cue pattern is a perfect index pattern, i.e. $\mathbf{z} = \mathbf{S}_\mu$, then the postsynaptic output can be written as:
\begin{equation}
    \mathbf{\hat{P}}_\mu = \frac{1}{K} \mathbf{W}^H \mathbf{S}_\mu = \frac{1}{K} \mathbf{P} \; \mathbf{S}^{*\top} \mathbf{S}  \mathbf{e}^\mu = \mathbf{P} \mathbf{e}^\mu + \mathbf{P} [\frac{1}{K}  \mathbf{S}^{*\top} \mathbf{S} - \mathbf{I}]  \mathbf{e}^\mu
\label{hetero_retrieval}
\end{equation}
where $\mathbf{e}^\mu$ is the $\mu$-th cardinal basis vector of $\mathbb{R}^M$. 
The RHS of [\ref{hetero_retrieval}] contains the signal and the noise term. The noise term is zero if the index vectors are exactly orthogonal. Random index patterns will have Gaussian interference noise, and readout will have a signal-to-noise ratio of approximately $2N/M$ \citep{Frady2018}. 
The crosstalk noise is complex-valued and will prevent [\ref{hetero_retrieval}] to produce a purely real-valued pattern.

\subsection*{Relating TPAM networks to spiking neural networks}

\subsubsection*{Details of spiking model}
\label{sec:spiking_methods}

Simulations were carried out using the Brian 2 simulator for spiking neural networks \citep{Goodman2009}. The dynamics of the integrate-and-fire neurons is given by:
\begin{equation}
C \frac{dV_{i}}{dt}= g_l (E_l - V_{i}) + I_{i} 
\label{eqn:if_dyn}
\end{equation}
where $C$ is the capacitance, $g_l$ is the leak conductance and $E_l$ is the resting potential. When the voltage of the neuron exceeds threshold $V_\theta$ then the neuron fires a spike and is reset to the reset potential $V_r$. 

Each pre-synaptic spike causes a synaptic state variable to increment. The synaptic state variable instantly jumps up after the delay time $\Delta_{ij} = \phi(W_{ij}) / (2 \pi \omega) + nT$, and decays exponentially:
\begin{equation}
    \frac{d s_{ij}}{dt} = - \frac{s_{ij}}{\tau_{s}} + \sum_k \delta \left( t - \Delta_{ij} - t_{j}^{(k)}  \right)
\end{equation}
where $t_j^{(k)}$ indicates the timing of the pre-synaptic spikes.

Each synapse contributes current to the post-synaptic cell proportional to its state-variable $s$:
\begin{equation}
I_{i} = \sum_j |W_{ij}| s_{ij}
\end{equation}

The excitatory currents are balanced by inhibitory currents, which are routed through a population of inhibitory neurons. The timeconstants of the synapses are set to $T$ for inhibitory and $0.5T$ for excitatory, and the current recombines to create an oscillation with cycle period of $T$. The filtering of the neural capacitance also contributes to shaping the synaptic inputs into a sine wave. The neural timeconstant is $0.25T$ for excitatory neurons and $0.1T$ for inhibitory neurons. These are relatively small and their contributions to the overall dynamics can be generally ignored. The timeconstants can be tuned to better approximate a perfect oscillation. 

The gain of the inhibition can be controlled through several mechanisms. Scaling the synaptic weights, which can be the E-to-I or I-to-E synapses, the size of the inhibitory population, and the timeconstants of the synaptic connection each contribute a proportional factor to the overall gain of the inhibition. The final factor is accounted by the gain of the neural transfer function, which can be approximated as follows. 

The integrate-and-fire neuron has an analytically defined approximation of the neural transfer function (for fixed/slow input), which we utilize to set the parameters of the spiking model. Based on a constant current into the neuron, one can compute the time it takes to integrate from the reset potential to the threshold potential:
\begin{equation}
T_{spike} = \frac{1}{g} \log \left( \frac{I - V_{r}}{I - V_{\theta}} \right) + \tau_{ref}
\end{equation}
where $g$, $V_r$, $V_\theta$ and $\tau_{ref}$ are parameters of the integrate-and-fire neuron model. The `current' $I$ is the input value, directly related to the dendritic sum variable $u_i$ in the normative TPAM network.

The \emph{instantaneous firing rate} ($IFR$) is used to map the neural transfer function to the spiking neurons. It is the inverse of the spiking period: $IFR=1/T_{spike}$. If there is no refractory period ($\tau_{ref}=0$), then the $IFR$ will asymptotically converge to a straight line for large input currents
$\tilde{IFR} = m I + b$, with
\begin{align}
\begin{split}
    m &= g / (V_\theta - V_{r})\\
    b &= -\frac{g (V_\theta + V_{r})}{2(V_\theta - V_{r})}
    \end{split}
\label{eqn:ifr_lin_approx}
\end{align}

With a refractory period, the $IFR$ will have similar properties, but will saturate at $1/\tau_{ref}$. 

The refractory period is used to change the non-linear transfer function of the neuron. For neurons as part of a TPAM network, with a phasor-projection as the non-linearity, a large refractory period that limits the neurons to one spike per cycle can mimic the phasor-projection. Typically, we use something like $0.6T$.

For the inhibitory population and the readout neurons, which act linearly with input current, we set the parameters so that the $IFR$ is a linear function of the input, by using zero (or very small) refractory period. The gain of the neural transfer function is then estimated by the slope of the IFR  [\ref{eqn:ifr_lin_approx}].

The parameters of the network are chosen to approximate biological parameters. The capacitance of the neurons does affect the dynamics of the network, and can itself act as a slight fixed delay. However, any such delay due to timeconstant could be compensated for by reducing the synaptic conduction delays.
The jump discontinuity in the synaptic dynamics could be modeled with a rise-time, which can be better tuned to produce a more ideal postsynaptic currents.
With careful consideration, the more detailed parameters of the spiking model could be better accounted for. However, the parameters do not have to be perfect to get a model working. 

\subsubsection*{Spiking capacity experiments}
\label{sec:spike_capacity_methods}

To be sure that the spiking model approximations did not catastrophically interfere with the attractor dynamics, we ran similar capacity experiments on the spiking model as the normative model, but at a much smaller scale. We chose a fixed parameter set of $N=500$ neurons with $K=25$ active and stored $M=[50, 100, 200, 250, 300, 400]$ random phasor patterns in the network. These parameters were not the optimized parameters. 

The spiking network was initialized to one of the stored patterns and iterated for 5 seconds, which is about 25 cycles. We measured the spike timings in the last second of the simulation and computed the exact frequency of the spiking oscillation. This was then translated into a complex vector. The complex phase of each element in this vector was then rotated to best align with the target pattern. The similarity is then the normalized dot product between the aligned vector and the target pattern.

The details of the spiking model currently prevent an exact parameter match between the normative and the spiking model, but we do see that a consistent spiking model follows the same capacity trajectory as a normative model. Without much parameter optimization, we can build a spiking attractor network that can easily store more patterns than a traditional Hopfield model.

\bibliographystyle{ieeetr}
\bibliography{hdspike}

\end{document}